\documentclass[a4paper,12pt]{article}

\usepackage{import}
\usepackage{color}
\usepackage{url}

\usepackage{graphicx}

\usepackage[cmex10]{amsmath}
\usepackage{amssymb}
\usepackage{array}

\begin{document}
\title{\textbf{Robotic Arm for Remote Surgery}}

\author{Steven~Dinger%
\thanks{S. Dinger  is with the Biomedical Engineering Research Group, School of Electrical \& Information Engineering, University of Witwatersrand, Johannesburg,
 e-mail: steven.dinger@wits.ac.za.}
        , John~Dickens and Adam Pantanowitz% stops a space
% <-this % stops a space
\thanks{J. Dickens and A. Pantanowitz are with the Biomedical Engineering Research Group, School of Electrical \& Information Engineering, University of Witwatersrand, Johannesburg.}}% <-this % stops a space
\date{}


\markboth{Journal of Medical Engineering, 2012}%
{Shell \MakeLowercase{\textit{et al.}}: Journal of Medical Engineering}



\maketitle

\begin{abstract}
\noindent{R}ecent advances in telecommunications have enabled surgeons to operate remotely on patients with the use of robotics.
The investigation and testing of remote surgery using a robotic arm is presented. The robotic arm is designed to have four degrees
of freedom that track the {surgeon's} $x,~y,~z$ positions and the rotation angle of the forearm $\theta$. The system comprises two main subsystems \textit{viz}. the detecting and actuating systems. The detection system uses infrared light-emitting diodes, a retroreflective bracelet and two infrared cameras which as a whole determine the coordinates of the {surgeon's} forearm. The actuation system, or robotic arm, is based on a lead screw mechanism which can obtain a maximum speed of $0.28~m.s^{-1}$ with a $1.5~^{\circ}.step^{-1}$ for the end-effector. The infrared detection and encoder resolutions are below $0.6~mm.pixel^{-1}$ and $0.4~mm$ respectively, which ensures the robotic arm can operate precisely. The surgeon is able to monitor the patient with the use of a graphical user interface on the display computer. The lead screw system is modelled and compared to experimentation results. The system is controlled using a simple proportional-integrator (PI) control scheme which is implemented on a dSpace control unit. The control design results in a rise time of less than $0.5~s$, a steady-state error of less than $1~mm$ and settling time of less than $1.4~s$. The system accumulates, over an extended period of time, an error of approximately $4~mm$ due to inertial effects of the robotic arm. The results show promising system performance characteristics for a relatively inexpensive solution to a relatively advanced application.\vspace{5mm}

\noindent{K}eywords: Remote surgery, PI controller, lead screw, optical encoder, infrared light emitting diode.
\end{abstract}

%\begin{IEEEkeywords}

%\end{IEEEkeywords}

% \fi
%\IEEEpeerreviewmaketitle

\section{Background}
Remote surgery or telesurgery enables a surgeon to operate on a patient who is at a different geographical
location. The use of robotics in surgery has improved the accuracy and capabilities of surgeon's significantly in recent years~\cite{GB:2006}. The major drawback, however, with current systems is that the surgeon is usually required to control the robotic device in an uncomfortable position with the device usually vulnerable to human tremors and operator fatigue~\cite{Dholt:2004}.

The concept of remote surgery is explored and tested with the use of a robotic arm that is able to mimic the movement of a surgeon's forearm. The robotic arm is designed to have four degrees of freedom \textit{viz}. the $x,~y,~z$ positions and rotation angle $\theta$ of the forearm. The simplest requirement of the robotic arm is to track the movements of a surgeon's forearm precisely and to operate in an identical workspace to that of a typical surgeon.

The three types of human movement detection systems are the inside-in, outside-in and inside-out detection systems~\cite{AM:1994}. The use of an accelerometer to detect three-dimensional movement is a feasible technology and represents an inside-in type system~\cite{YT:2006}. The use of potentiometers, piezo-resistive flex and cable extension represent examples of outside-in type systems~\cite{AM:1994}, however they suffer from  drift.

A spherical design of the robotic arm overcomes many of the problems faced in robotics, such as reducing the size and weight of the robotic instrument~\cite{JR:2005}. The reachable workspace, however, for a spherical robotic arm is limited to a sector of a sphere, and is therefore not desirable in terms of reproducing the complete operating workspace of the surgeon.

The telesurgery robotic arm comprises two major non-collocated subsystems, \textit{viz}. the detection and actuation system. The coordination between the two systems can be achieved through the use of current telecommunications technology~\cite{GB:2006}.

This paper specifies the system requirements, constraints and assumptions that are used to effectively implement such a system. The layout of the system is presented and covers the two main subsystems, \textit{viz}. the detection system and the actuation system. The lead screw system is modelled and used to compare to the experimentation results. The controller used for the robotic arm is also discussed. The simulation and experimentation results are provided and analysed in order to determine the performance of the system. The system limitations are listed including recommendations and suggestions for future work. A conclusive summary pertaining to the performance and fundamental aspects of the system is given.


% \hfill January 11, 2007
\section{System Specifications}\label{sec:req}
\noindent{T}he forearm of a surgeon is considered to have four degrees of freedom, which include the $x,~y,~z$ positions
and rotation of the forearm denoted as $\theta$. A satisfactory design, therefore, includes the ability of the
robotic arm to mimic and track the surgeon's forearm for all four degrees of freedom.

The delay of the control signal of the system must be less than $500~ms$ for the surgeon to successfully compensate for the latency during the operating procedure~\cite{BC:2003}. Visual feedback is required in order for the surgeon to monitor the patient and correct for any error that accumulates during the procedure. The tracking error is considered more important as opposed to the steady-state error, since the steady-state error can be easily corrected by the surgeon through the visual feedback system. The system must be able to be calibrated to remove any cumulative error that occurs during operation.

A constraint on the amount of wrist pitch and forearm flexion of the surgeon exists mainly due to the outside-in type of detection system employed~\cite{AM:1994}. The outside-in detection system uses artificial sources placed on the body with the sensors situated externally.

\subsection{Assumptions}
\noindent{T}he assumptions used to design, implement and operate the system successfully are:

\begin{itemize}
 \item Pitch of the surgeon's hand is negligible during an operation.
 \item A maximum speed of $0.25~m.s^{-1}$ for each axis is deemed satisfactory.
 \item The workspace of the surgeon is devoid of infrared noise.
\end{itemize}

\noindent{T}he assumption that the hand pitch is negligible is necessary, since the detection sources are placed on the hand as described in Section~\ref{sec:det}. A maximum speed of $0.25~m.s^{-1}$ is reasonable, since a surgeon's forearm
should remain in the same anatomical region throughout the operation, with the hand performing most of the fast and precise movements. A speed of $0.25~m.s^{-1}$, which was measured experimentally, is also relatively large for a typical human forearm to move. The workspace should be illuminated with fluorescent lighting in order to reduce infrared noise.

\section{System Architecture}
\begin{figure}[htb!]
	\centering
	\def\svgwidth{\columnwidth}
        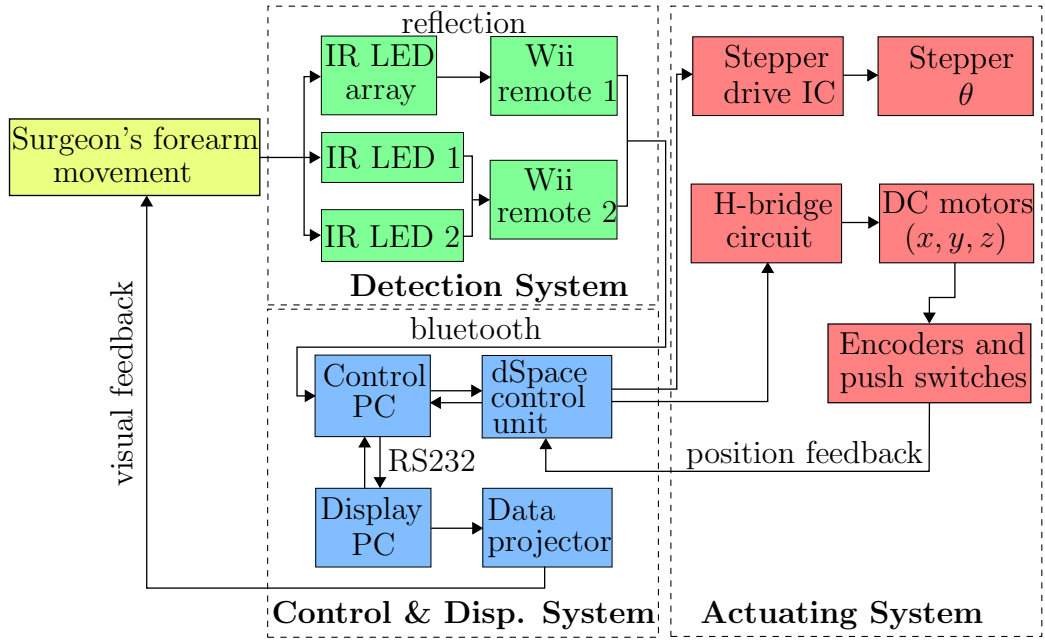
        \caption{System block diagram.}
        \label{fig:sys}
\end{figure}

\noindent{T}he system comprises two main subsystems \textit{viz}. the detection and actuation systems. The
block diagram that illustrates how the two systems are coordinated and linked as shown in Figure~\ref{fig:sys}.

\subsection{Detection System}\label{sec:det}
\noindent{T}he movement of the surgeon's forearm is detected using a non-invasive outside-in detection system, which consists of placing the sensors off, and artificial sources on, the person. The artificial sources used are two infrared ($940~nm$) light emitting diodes (LEDs) and a retroreflective bracelet.

\begin{figure}[htb!]
	\centering
	\def\svgwidth{\columnwidth}
        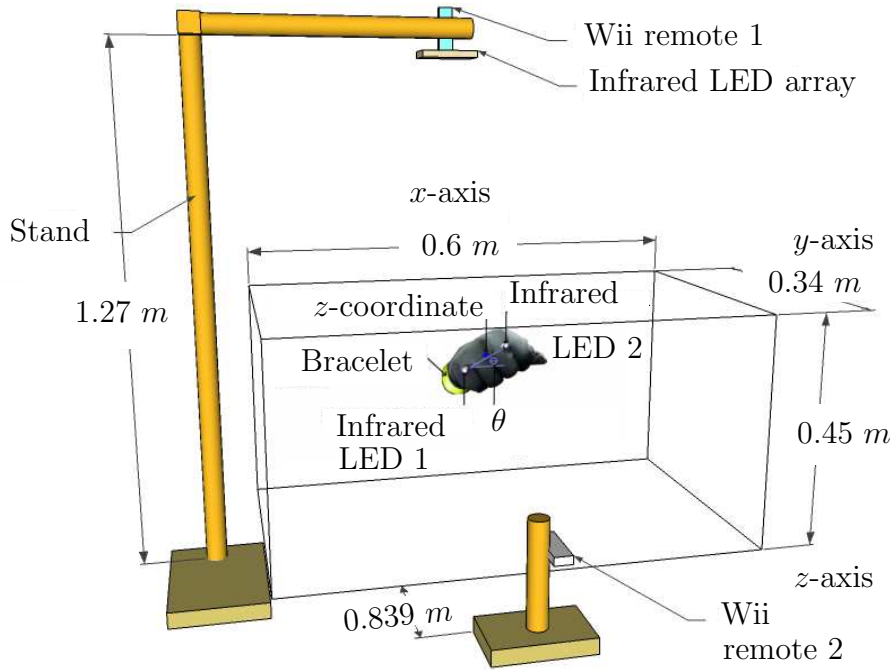
        \caption{Illustration of the infrared detection system.}
        \label{fig:WiiP}
\end{figure}

The two infrared LEDs are located in the region of the proximal phalanges and are used to measure the
$z$-axis position of the surgeon's arm, including the forearm rotation angle $\theta$, as shown in Figure~\ref{fig:WiiP}. Two Nintendo Wii remotes are used to capture and process the data obtained from the infrared sources and reflections. The Wii remote infrared ($940~nm$) camera has a resolution of $1024\times768$, which at the heights and distances shown in Figure~\ref{fig:WiiP}, provides a resolution of $0.586~mm.pixel^{-1}$ for $x$ and $z$ axes, and $0.576~mm.pixel^{-1}$ for the $y$ axis. A retroreflective wrist band is used as opposed to infrared LEDs in order for the Wii remote to continue tracking the forearm under rotation. A 132-LED array is used to produce the infrared light source for the reflective wrist band~\cite{AP:2008}.

\subsection{Actuation system}
\begin{figure}[htb!]
	\centering
	\def\svgwidth{\columnwidth}
        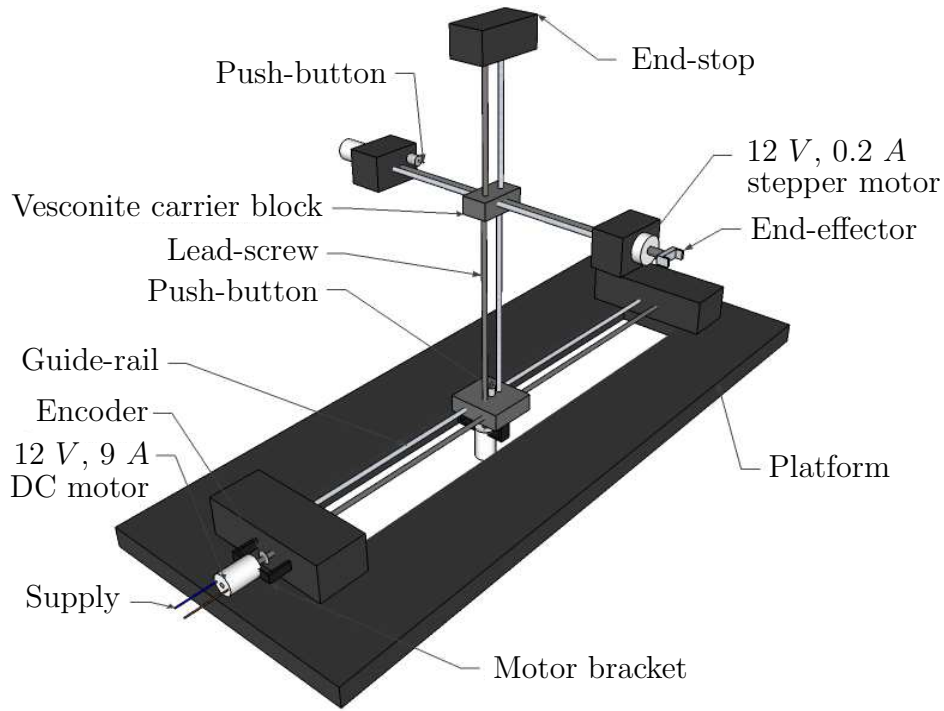
        \caption{Illustration of the mechanical robotic arm.}
        \label{fig:Mech}
\end{figure}

\noindent{T}he actuation system is illustrated in Figure~\ref{fig:Mech}, and includes the optical encoders that measure the position of the carrier blocks and calibration switches located on each axis. The robotic arm follows a simple lead-screw guide-rail design, with the advantages of the design being:
\begin{itemize}
 \item No energy consumption at rest.
 \item Stable and rigid.
 \item Lightweight and modular.
 \item Simple to construct and control.
\end{itemize}

\noindent{T}he disadvantages of the design include a low mechanical efficiency due to excessive amounts of dynamic
friction~\cite{KW:2006}. The non-linear static friction in the system causes a deadzone effect, whereby the control signal has no effect on the system within a certain armature voltage range. The fine standard pitch used for the lead screws is $1.5~mm$ for the $x$ and $z$ axes, and $1.25~mm$ for the $y$ axis. The size of pitch requires large angular speeds ($17000~rpm$) to generate modest linear speeds. The large angular speeds cause mechanical vibrations, an increase in friction (heat losses) and significant audible noise.
\section{Control}
\subsection{System Interfacing}
\noindent{T}he detected infrared information is processed and conditioned as inputs for the $dSpace~1104$ control unit. The desired coordinates are written to dSpace using the commands and functions supplied by the Matlab library called $mlib$.

The dSpace unit interfaces with the actuation system and communicates with the control computer. The model of the controller is built in the Matlab Simulink environment, including the necessary routing such as the inputs from the encoders and outputs to the logic and pulse-width modulation~(PWM) channels.

The $1.5~^{\circ}.step^{-1}$ geared stepper motor has no feedback and is rotated using open-loop, single-phase excitation \cite{PC:1997}\cite{SL:2009}. A step frequency of $125~Hz$ is used, which results in an angular speed of $31.25~rpm$.

\subsection{System Modeling}\label{app:model}
The model of the system is presented and is used to obtain the gain values for the chosen proportional-integral~(PI) controller. The assumptions used to model the system are listed below:
\begin{itemize}
	\item Inelastic steel lead screw.
	\item Motor inductance is negligible.
	\item Static friction coefficient for lubricated metals.
	\item The lead screw moment of inertia is larger than that of the motor shaft.
	\item Mechanical power loss in the motor is negligible.
\end{itemize}

\noindent{T}he assumption that elastic torsion of the threaded steel rod does not occur is validated since the rods are short in the lengths~(less than $1~m$). The mass accelerated by the motors is much larger than the inertia introduced by the inductance of the motor. The mass of the lead screw is much larger than that of the motor shaft, as shown in equation~\ref{eqn:MassLead}, which validates the assumption that the lead screw moment of inertia is greater than that of the motor shaft. The mechanical windage and frictional losses in the motor are assumed to be negligible in order to approximate the back-EMF constant $K_{b}$ to the armature torque constant $K_{a}$, and is shown in detail by Sen~\cite{PC:1997}.

\begin{figure}[htb!]
	\centering
	\def\svgwidth{\columnwidth}
	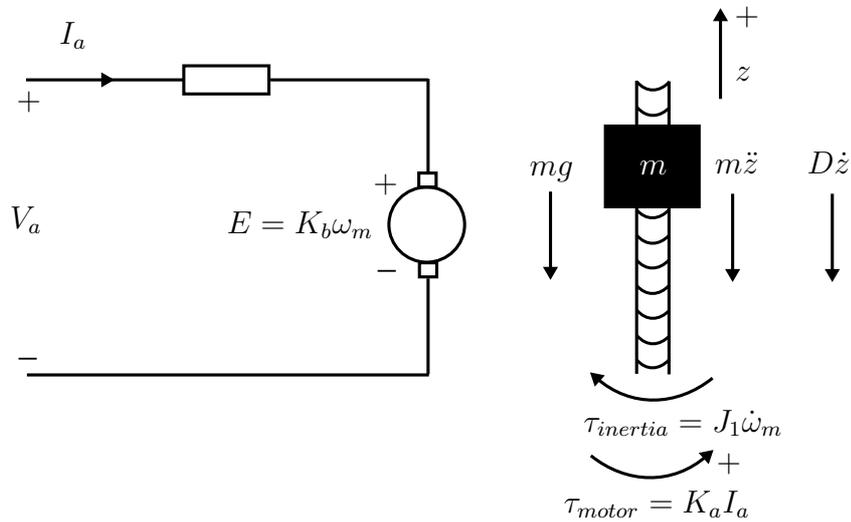
	\caption{System model diagram.}
\label{fig:free}
\end{figure}

The dynamics of the vertical~($z$-axis) lead screw are obtained with the gravitational acceleration~($g~=~9.8~m/s^{2}$) treated as a constant disturbance on the system. The system model diagram, consisting of the DC motor and lead screw, is shown in Figure~\ref{fig:free}. The system model diagram is used to derive the system equations by using Newton's laws of motion, with the relationship between motor torque and load weight provided by Hollander and Sugar~\cite{KW:2006}. The static friction coefficient is assumed to have a value of 0.06, which is the value for lubricated metals~\cite{KW:2006}.
\newpage
\noindent{T}he definitions of the parameters and quantities used to describe the lead screw and motor system are given as:
\begin{center}
\begin{tabular}{l}
$m$~~-~~Load~mass~($kg$)\\[1.6mm]
$D$~~-~~Dynamic~friction~coefficient~($N.s/m$)\\[1.6mm]
$\mu$~~-~~Static~friction~coefficient\\[1.6mm]
$r$~~-~~Lead~screw~radius~($m$)\\[1.6mm]
$L$~~-~~Lead screw length~($m$)\\[1.6mm]
$l$~~-~~Pitch~($m$)~($m/rev$)\\[1.6mm]
$K$~~-~~Pitch~constant~($m/rad$)\\[1.6mm]
$\tau_{motor}$~~-~~Motor~torque~($N.m$)\\[1.6mm]
$\tau_{inertia}$~~-~~Lead~screw~inertial~torque~($N.m$)\\[1.6mm]
$\tau_{mech}$~~-~~Resultant~mechanical~torque~($N.m$)\\[1.6mm]
$\omega_{m}$~~-~~Motor~angular~velocity~($rad/s$)\\[1.6mm]
$R_{a}$~~-~~Armature~resistance~($\Omega$)\\[1.6mm]
$J_{l}$~~-~~Lead~screw~moment~of~inertia~($kg.m^{2}$)\\
\end{tabular}
\end{center}

\noindent{T}he equation that describes the dynamics of the motor is obtained by using Kirchoff's voltage law, as shown in equation~\ref{eqn:kvl}.
\begin{eqnarray}
V_{a} &=& I_{a}R_{a} + K_{b}\omega_{m}\label{eqn:kvl}\\
\tau_{mech} &=& rF_{w}\frac{sin{\alpha}+\mu{cos{\alpha}}}{cos{\alpha}-\mu{sin{\alpha}}}\nonumber\\
\alpha &=& arctan\left(\frac{l}{2\pi{r}}\right)\nonumber\\
l &=& 1.25~mm\nonumber\\
r &=& 4~mm\nonumber\\
\therefore\alpha &=& 2.847~^{\circ}\nonumber\\
\Rightarrow \tau_{mech} &=& 0.4403\times10^{-3}\times{F_{w}}\nonumber\\
\end{eqnarray}
Let:
\begin{eqnarray}
\gamma &=& 0.4403\times10^{-3}\nonumber\\
\end{eqnarray}
Define:
\begin{eqnarray}
\tau_{mech} &=& \tau_{motor}-\tau_{inertia}\nonumber
\end{eqnarray}

\noindent{T}he resultant force $F_{w}$ is the sum of the opposing forces acting on the mass for motion in the positive $z$-direction. The equation that describes the relationship between the torques and opposing forces is shown in equation~\ref{eqn:torq}.
\begin{eqnarray}
F_{w} &=& D\dot{z}+mg+m\ddot{z}\label{eqn:res}\\
\tau_{motor}-\tau_{inertia} &=& \gamma\left(D\dot{z}+mg+m\ddot{z}\right)\label{eqn:torq}\\
\dot{z} &=& K\omega_{m}\nonumber
\end{eqnarray}

\noindent{T}he mass accelerated by the vertical lead screw, $m$, is approximately $1.3~kg$. The armature resistance of the brushed DC motor is provided by the data sheet with a value of $0.205~\Omega$~\cite{FL:2009}. The armature torque constant $K_{a}$ is approximately $10.25\times10^{-3}~N.m/A$~\cite{FL:2009}. The back-EMF constant $K_{b}$ is the same as the armature constant with a value of $10.25\times10^{-3}~V.s/rad$. The pitch per radian of rotation $K$ for the standard $8~mm$ diameter lead screw is calculated as $\frac{1.25\times10^{-3}}{2\pi}$, which is equal to $0.1989\times10^{-3}~m/rad$.

\begin{eqnarray}
\tau_{inertia} &=& J_{l}\frac{d\omega_{m}}{dt}\nonumber\\
\tau_{inertia} &=& \frac{J_{l}}{K}\ddot{z}\nonumber\\
J_{l} &=& \frac{1}{2}Mr^{2}\nonumber\\
M &=& {\pi}r^{2}L\rho_{steel}\nonumber\\
\rho_{steel} &=& 7850~{kg/m^{3}}\nonumber\\
L &=& 0.54~m\nonumber\\
\therefore M &=& 0.213~kg\label{eqn:MassLead}\\
\Rightarrow J_{l} &=& 1.71\times10^{-6}~kg.m^{2}\nonumber\\
\therefore \frac{J_{l}}{K} &=& 8.57\times10^{-3}~kg.m.rad\nonumber
\end{eqnarray}

\noindent{T}he two separate systems, motor and lead screw, are combined into one system with the appropriate substitutions shown in equation~\ref{eqn:kvll}.
\begin{eqnarray}
V_{a} &=& \frac{\tau_{motor}}{K_{a}}R_{a}+K_{b}\frac{\dot{z}}{K}\label{eqn:kvll}\\
V_{a} &=& \frac{R_{a}\gamma}{K_{a}}\left(D\dot{z}+mg+m\ddot{z}\right)+K_{b}\frac{\dot{z}}{K}+\frac{R_{a}}{K_{a}}\tau_{inertia}\nonumber\\
\therefore V_{a} &=& \left(\frac{R_{a}J_{l}}{K_{a}K}+\frac{R_{a}m\gamma}{K_{a}}\right)\ddot{z}+\left(\frac{DR_{a}\gamma}{K_{a}}+\frac{K_{b}}{K}\right)\dot{z}+\frac{R_{a}\gamma{mg}}{K_{a}}\label{eqn:diff}
\end{eqnarray}

\noindent{T}he dynamic friction coefficient $D$ is determined from the experimentation results by estimating the constant maximum velocity~($0.262~m/s$), which occurs at maximum voltage of $13.8~V$, and substituting the two values into equation~\ref{eqn:torq}~and~equation~\ref{eqn:kvll}. The dynamic friction value is calculated to be approximately $81~N.s/m$, $80~N.s/m$ and $575.6~N.s/m$ for the $z$, $x$ and $y$-axis lead screws respectively. The dynamic friction coefficient for the $y$-axis lead screw is significantly larger than the other two lead screws since it is driven by a different motor that has a fan attached to the motor shaft~\cite{JR:2009}. The simplified state-space representation of the system is obtained by substituting the listed variables and constants into equation~\ref{eqn:diff}.

\begin{center}
\begin{tabular}{ll}
$z~~=~~x_{1}$&$,~V_{a}~~=~~u_{1}$\\[1.6mm]
$\dot{z}~~=~~x_{2}~~=~~\dot{x_{1}}$&$,~g~~=~~u_{2}$\\[1.6mm]
$\ddot{z}~~=~~\dot{x_{2}}~~=~~\ddot{x_{1}}$&$,~y_{1}~~=~~x_{1}$\\[1.6mm]
$K_{1}~~=~~\left(\frac{DR_{a}\gamma}{K_{a}}+\frac{K_{b}}{K}\right)$&$,~y_{2}~~=~~x_{2}$\\[3mm]
$K_{2}~~=~~\left(\frac{R_{a}J_{l}}{K_{a}K}+\frac{R_{a}m\gamma}{K_{a}}\right)$&$,~K_{3}~~=~~\frac{R_{a}\gamma{m}}{K_{a}}$
\end{tabular}
\end{center}
The state equations are given as:
\begin{eqnarray}
\left[
\begin{array}{c}      
\dot{x_{1}}\\       
\dot{x_{2}} 
\end{array}
\right] &=& \left[
\begin{array}{cc}      
0&1\\       
0&\frac{-K_{1}}{K_{2}}
\end{array}
\right]{\left[\begin{array}{c}
x_{1}\\
x_{2}
\end{array}\right]}+\left[
\begin{array}{cc}
0&0\\
\frac{1}{K_{2}}&\frac{-K_{3}}{K_{2}}\end{array}
\right]\left[
\begin{array}{c}
u_{1}\\
u_{2}\end{array}
\right]\nonumber\\[3mm]
\left[\begin{array}{c}
y_{1}\\
y_{2}\end{array}\right] &=& \left[\begin{array}{cc}
1&1\end{array}\right]\left[
\begin{array}{c}
x_{1}\\
x_{2}\end{array}\right]\nonumber
\end{eqnarray}

\noindent{T}he non-linear friction in the system causes a deadzone effect to occur whereby the input control voltage is not able to move the mass until it is greater than a certain value. The value of the input voltage at which movement occurs was found through experimentation to be approximately $5~V$, $3~V$ and $3.4~V$ for the $z$, $x$ and $y$-axis lead screws respectively. The deadzone effect is included in the Simulink s-function, which is used to model the system. The derived state equations hold for the other two axes by equating the gravitational acceleration input to zero and substituting in the appropriate masses, motor constants and other adjusted parameters.

\subsection{Controller design}
\noindent{A} classic controller is used for the robotic arm and comprises a proportional gain and integrator, with gain constants of $K_{i}=150~V.m^{-1}.s^{-1}$ and $K_{p}=55~V.m^{-1}$. The gain constants were determined from the model and refined through experimentation. The integrator is limited in order to prevent integrator-wind up~\cite{RS:2001}. All of the axes are controlled with identical controllers.

A derivative term was not used since it was found from experimentation that it amplified noise and increased the motor armature current significantly~($\Delta{I_{a}}=3~A$). The controller was designed to optimize the actuation system in terms of power consumption, tracking error and rise time. The reduction of power consumption is achieved by switching the PWM off when the position error is within a $1~mm$ error band for a duration of $5~s$. The trade-off, however, is that there exists some steady-state and overshoot error in the step response.

\section{System Performance and Analysis}
\begin{figure}[htb!]
	\centering
	\def\svgwidth{\columnwidth}
        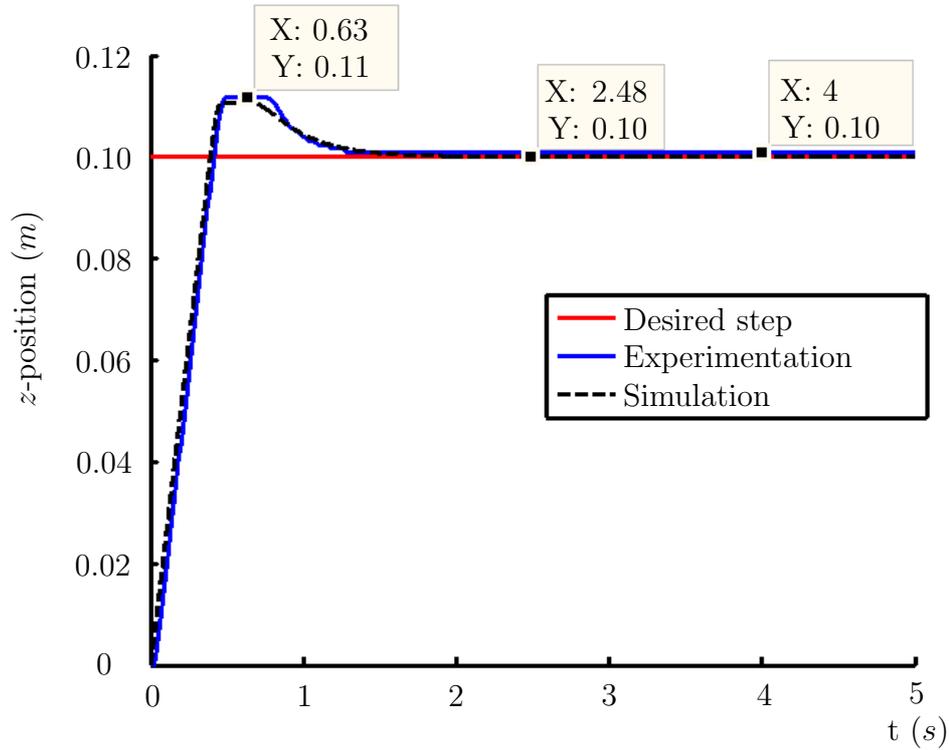
        \caption{Step response of the $z$-axis lead screw.}
        \label{fig:zstep}
\end{figure}

\begin{figure}[htb!]
	\centering
	\def\svgwidth{\columnwidth}
        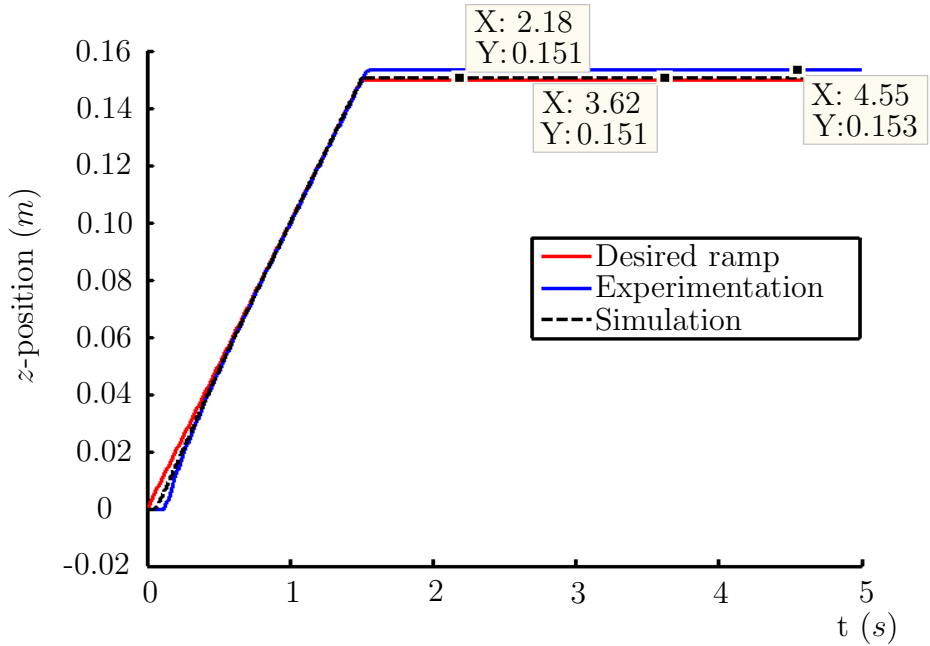
        \caption{Ramp response of the $z$-axis lead screw.}
        \label{fig:zramp}
\end{figure}

\begin{figure}[htb!]
	\centering
	\def\svgwidth{\columnwidth}
        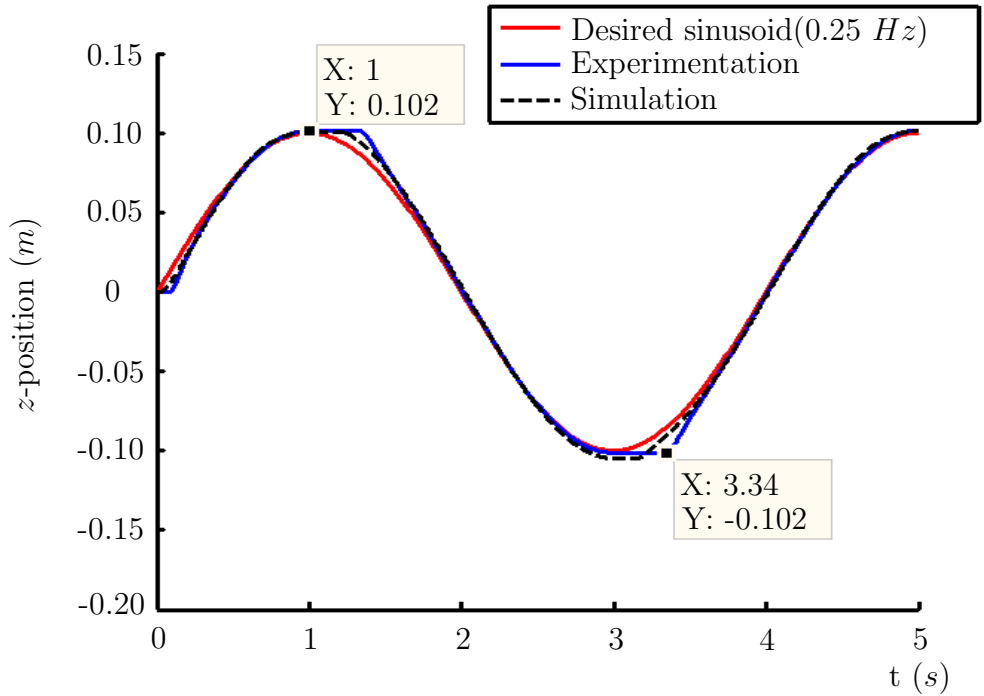
        \caption{Sinusoidal response of the $z$-axis lead screw.}
        \label{fig:zsin}
\end{figure}

\noindent{T}he system performance was measured using sinusoid, ramp and step inputs. The rise time and tracking
error are obtained from the step response and ramp response respectively. The steady-state error and settling time are also obtained from the step response of the system. The deadzone time is the duration that the system remains stationary due to non-linear friction, and is most evident in the sinusoidal response. The reliability of the system was measured by observing the accumulated error for a specified time limit. These results were compared to simulations based on the Newtonian models. The various system responses for the $z$-axis lead screw are shown in Figures~\ref{fig:zstep},~\ref{fig:zramp} and~\ref{fig:zsin}. The results for the other two axes are similar.

The overall results of the system performance in terms of responsiveness and resolutions are shown in Table~\ref{tab:results}. The encoder resolution is determined by dividing the lead screw pitch by the number of opaque sectors~(4 dark segments per axis encoder).
\newpage

\renewcommand{\arraystretch}{1.5}
\begin{table}[htb!]
    \caption{Experimentation results of system performance.\label{tab:results}}
    \begin{center}
        \begin{tabular}{|l|c|c|c|}
        \hline
         {\textbf{Parameter}}&{\textbf{\mbox{\boldmath$x$}-axis}}&{\textbf{\mbox{\boldmath$y$}-axis}}&{\textbf{\mbox{\boldmath$z$}-axis}}\\
        \hline
         Steady-State Error~($mm$)&0.9&0.313&0.9\\
        \hline
         Rise Time~($s$)&0.35&0.28&0.41\\
        \hline
         Overshoot~($\%$)&16&16.3&11.9\\
        \hline
         $2~\%$~Settling Time~($s$)&1.31&0.95&1.21\\
        \hline
         Deadzone Time~($s$)&0.33&0.35&0.33\\
        \hline
         Encoder Resolution~($mm$)&0.375&0.313&0.313\\
        \hline
         Resolution~($mm.pixel^{-1}$)&0.586&0.586&0.576\\
        \hline
        \end{tabular}
    \end{center}
\end{table}
\newpage
\noindent{T}he results show that the system is able to perform satisfactorily according to the success criteria stated in Section~\ref{sec:req}. It is noted that the rise time of $y$-axis lead screw is the fastest, which is due to the fact that the motor has a higher rated speed compared to the other two DC motors~\cite{JR:2009}\cite{FL:2009}.

The error accumulation in the system was found to be a function of speed and sudden changes in motor direction. The cumulative error was therefore quantified using two different waveforms: a low-frequency sinusoid and a higher harmonic triangular waveform. The sinusoid and triangular waveforms were each applied separately for a duration $68~s$ and $41~s$ respectively. The offset error was recorded at the end of each test with the average results shown in Table~\ref{tab:error}.

The cumulative error is due to the inertia of the masses opposing sudden changes in velocity, and since the motor direction switches electronically, results in an increment or decrement of the position. The error therefore accumulates when there are sudden changes in direction at a reasonable speed. An error is also associated with the incremental measurement made by the low-cost optical encoders, which can be corrected by using alternative encoders. The error accumulation is largest as expected for the triangular waveform, as shown in Table~\ref{tab:error}.

\begin{table}[htb!]
    \caption{Experimentation results of system error accumulation.\label{tab:error}}
    \begin{center}
        \begin{tabular}{|l|c|c|c|}
        \hline
        \textbf{Parameter}&\multicolumn{3}{|c|}{\textbf{Axis}}\\
        \cline{2-4}
         ~&\mbox{\boldmath$x$}&\mbox{\boldmath$y$}&\mbox{\boldmath$z$}\\
        \hline
         Waveform~($m$)&\multicolumn{3}{|c|}{$0.1\times{sin(2\pi(0.25)t)}$}\\
        \hline
         Offset Error~($mm$)&0.938&3.73&2.5\\
        \hline
         Displacement~($m$)&6.8&6.8&6.8\\
        \hline
         Error~($\%$)& 0.0138&0.0549&0.0368\\
        \hline
        Waveform~($m$)&\multicolumn{3}{|c|}{$0.1\times{tri(2\pi(0.375)t)}$}\\
        \hline
         Offset Error~($mm$)&11.6&7.34&24.1\\
        \hline
         Displacement~($m$)&4.4&4.3&4.4\\
        \hline
         Error~($\%$)& 0.264&0.171&0.548\\
         \hline
        \end{tabular}
    \end{center}
\end{table}

\newpage
\subsection{System limitations}
\noindent{T}he telesurgery robotic arm system has a number of limitations \textit{viz}.:

\begin{itemize}
\item Average maximum speed of $0.28~m/s$.
\item Reasonable mechanical vibrations and resonance.
\item Mechanical stability varies as a function of $z$-position.
\item Maximum instantaneous power consumption of $360~W$.
\item Detection is vulnerable to infrared noise.
\item Inertial effects cause error accumulation.
\item Controlled system exhibits overshoot and steady-state error.
\end{itemize}

\noindent{T}he majority of the limitations are due to the mechanical aspect of the system and can be improved with further design and development.
\newpage
\section{Conclusion}
\noindent{T}he objective of remote surgery is tested through the design and implementation of a relatively inexpensive robotic arm. The robotic arm consists of the detection system, which is implemented using the Nintendo Wii remote technology, and the actuation system which is implemented using a lead screw design. The robotic arm is able to track the surgeon's forearm with less than $1~mm$ steady-state error and with a rise time of less than $0.5~s$. The non-linear static friction, however, creates a deadzone effect that delays the motion of the robotic arm by $0.35~s$.

The detection and encoder resolutions are both submillimeter, which ensures that the robotic arm can move and operate precisely during surgery. The error accumulated in the system is below $4~mm$ when the robotic arm is operated smoothly and at slower speeds. The end-effector of the robotic arm could be further developed by including the {surgeon's} precise finger and hand movements measured using more sensors.

% or


% \appendices
% % you can choose not to have a title for an appendix
% % if you want by leaving the argument blank
% \section{}

\section*{Acknowledgment}
\noindent{T}he authors would like to thank Professor David Rubin and Mr Harold Fellows at the University of Witwatersrand for their guidance, advice and support.

%\ifCLASSOPTIONcaptionsoff
  %\newpage
%\fi



\bibliographystyle{IEEEtran}
\bibliography{IEEEabrv,arXiv}
\end{document}